\documentclass[sigconf]{acmart}
\def\BibTeX{{\rm B\kern-.05em{\sc i\kern-.025em b}\kern-.08emT\kern-.1667em\lower.7ex\hbox{E}\kern-.125emX}}
    
\usepackage{nicefrac}
\usepackage{siunitx}
\usepackage{array,framed}
\usepackage{booktabs}
\usepackage{
  color,
  float,
  epsfig,
  wrapfig,
  graphics,
  graphicx,
  subcaption
}
\usepackage{textcomp}
\usepackage{setspace}
\usepackage{latexsym,fancyhdr,url}
\usepackage{enumerate}
\usepackage{algorithm2e}
\usepackage{algpseudocode}
\usepackage{graphics}
\usepackage{xparse} 
\usepackage{xspace}
\usepackage{multirow}
\usepackage{csvsimple}
\usepackage{balance}
\usepackage{listings}
\usepackage{xcolor}
\usepackage{tabularx}
\usepackage{bm}
\usepackage{amsfonts}
\usepackage{xcolor}
\usepackage{diagbox}
\usepackage{booktabs}
\usepackage{multirow}
\usepackage{enumitem}

\usepackage{graphicx}

\usepackage{pifont}

\newcommand{\cmark}{\textcolor{green!50!black}{\ding{51}}}
\newcommand{\xmark}{\textcolor{red}{\ding{55}}}

\usepackage{adjustbox}

\usepackage{
  tikz,
  pgfplots,
  pgfplotstable
}
\usepackage{hyperref}

\usepackage{ulem} 

\usepackage{pifont}

\usetikzlibrary{
  shapes.geometric,
  arrows,
  external,
  pgfplots.groupplots,
  matrix
}

\pgfplotsset{compat=1.9}

\usepackage{mathtools,}

\DeclareMathAlphabet{\mathcal}{OMS}{cmsy}{m}{n}
\DeclareGraphicsExtensions{%
    .png,.PNG,%
    .pdf,.PDF,%
    .jpg,.mps,.jpeg,.jbig2,.jb2,.JPG,.JPEG,.JBIG2,.JB2}

\usepackage{xparse}
\newcommand{\bnm}{\begin{newmath}}
\newcommand{\enm}{\end{newmath}}

\newcommand{\bea}{\begin{eqnarray*}}%
\newcommand{\eea}{\end{eqnarray*}}%

\newcommand{\bne}{\begin{newequation}}
\newcommand{\ene}{\end{newequation}}

\newcommand{\bal}{\begin{newalign}}
\newcommand{\eal}{\end{newalign}}

\newenvironment{newalign}{\begin{align}%
\setlength{\abovedisplayskip}{4pt}%
\setlength{\belowdisplayskip}{4pt}%
\setlength{\abovedisplayshortskip}{6pt}%
\setlength{\belowdisplayshortskip}{6pt} }{\end{align}}

\newenvironment{newmath}{\begin{displaymath}%
\setlength{\abovedisplayskip}{4pt}%
\setlength{\belowdisplayskip}{4pt}%
\setlength{\abovedisplayshortskip}{6pt}%
\setlength{\belowdisplayshortskip}{6pt} }{\end{displaymath}}

\newenvironment{newequation}{\begin{equation}%
\setlength{\abovedisplayskip}{4pt}%
\setlength{\belowdisplayskip}{4pt}%
\setlength{\abovedisplayshortskip}{6pt}%
\setlength{\belowdisplayshortskip}{6pt} }{\end{equation}}

\newcounter{ctr}

\newcounter{mytable}
\def\mytable{\begin{centering}\refstepcounter{mytable}}
\def\endmytable{\end{centering}}

\newcounter{myfig}
\def\myfig{\begin{centering}\refstepcounter{myfig}}
\def\endmyfig{\end{centering}}

\newlength{\saveparindent}
\newlength{\saveparskip}
\newcommand{\E}{{\rm I\kern-.3em E}}

\renewcommand{\eqref}[1]{\mbox{Equation~(\ref{#1})}}

\def \part {part}

\renewcommand{\paragraph}[1]{\vspace*{6pt}\noindent\textbf{#1}\;}

\def \blackslug{\hbox{\hskip 1pt \vrule width 4pt height 8pt
    depth 1.5pt \hskip 1pt}}
\def \qed{\quad\blackslug\lower 8.5pt\null\par}

\newcounter{mynote}[section]

\newcommand\ignore[1]{}

\newcounter{rcnote}[section]

\newcounter{mrnote}[section]

\newcounter{fknote}[section]

\newcounter{anote}[section]

\DeclareMathSymbol{\mlq}{\mathord}{operators}{``}
\DeclareMathSymbol{\mrq}{\mathord}{operators}{`'}

\newcommand{\rhf}[2]{R_{f, \gamma}}

\DeclareDocumentCommand{\edist}{o o}{
  \ensuremath{
    \IfNoValueTF{#1}{{d}}{{\sf d}(#1,#2)}
  }
}

\newcommand{\olrk}[1]{\ifx\nursymbol#1\else\!\!\mskip4.5mu plus 0.5mu\left(\mskip0.5mu plus0.5mu #1\mskip1.5mu plus0.5mu \right)\fi}

\NewDocumentCommand{\indseq}{ O{1} O{r} }{{#1}\ldots {#2}}

\begin{document}
\fancyhead{}
\def\thetitle{Perturbation-efficient Zeroth-order Optimization for Hardware-friendly On-device Training}
\title{\thetitle}


\author{
Qitao Tan\textsuperscript{1}, Sung-En Chang\textsuperscript{2}, Rui Xia\textsuperscript{3}, Huidong Ji\textsuperscript{4}, Chence Yang\textsuperscript{1},Ci Zhang\textsuperscript{1}\\
Jun Liu\textsuperscript{2}, Zheng Zhan\textsuperscript{2}, Zhenman Fang\textsuperscript{5}, Zhuo Zou\textsuperscript{4}, Yanzhi Wang\textsuperscript{2}, Jin Lu\textsuperscript{1}
, Geng Yuan\textsuperscript{1}
}

\affiliation{%
\textsuperscript{1}University of Georgia \quad
\textsuperscript{2}Northeastern University \quad
\textsuperscript{3}University of Pennsylvania \\
\textsuperscript{4}Fudan University \quad
\textsuperscript{5}Simon Fraser University \quad
}

\begin{abstract}
    Zeroth-order (ZO) optimization is an emerging deep neural network (DNN) training paradigm that offers computational simplicity and memory savings. 
However, this seemingly promising approach faces a significant and long-ignored challenge. ZO requires generating a substantial number of Gaussian random numbers, which poses significant difficulties and even makes it infeasible for hardware platforms, such as FPGAs and ASICs.
In this paper, we identify this critical issue, which arises from the mismatch between algorithm and hardware designers. 
To address this issue, we proposed PeZO, a perturbation-efficient ZO framework. Specifically, we design random number reuse strategies to significantly reduce the demand for random number generation and introduce a hardware-friendly adaptive scaling method to replace the costly Gaussian distribution with a uniform distribution. Our experiments show that PeZO reduces the required LUTs and FFs for random number generation by 48.6\% and 12.7\%, and saves at maximum 86\% power consumption, all without compromising training performance, making ZO optimization feasible for on-device training. To the best of our knowledge, we are the first to explore the potential of on-device ZO optimization, providing valuable insights for future research.
\end{abstract}
\maketitle

\section{Introduction}

On-device training for deep neural networks (DNNs) has shown its irreplaceable role, as it enables the adaptiveness of deep learning applications to provide customized services while ensuring the protection of user data privacy. 
Although significant progress has been made in recent years for efficient inference~\cite{zeng2024flightllm,ma2019performance, kim2017fpga}, on-device training still remains a highly challenging problem.
The reason is that training requires significantly more memory compared to inference and involves various complex operations and data dependencies.

Recently, Zeroth-order (ZO) optimization has emerged as a memory-efficient training paradigm, attracting significant attention~\cite{zhang2024revisiting,liu2024sparse,malladi2023fine}.
This method intends only to use forward passes (i.e., inference) to estimate the gradients required for model training and update the weights accordingly.
By completely avoiding the computation of backward propagation, it eliminates the need to store the extensive activations and gradients and simplifies the computations.
As reported in MeZO~\cite{malladi2023fine}, fine-tuning language models (LMs) on a GPU server via ZO optimization reduces up to 12$\times$ memory cost and saves 2$\times$ computational cost per iteration.
ZO optimization addresses the two primary challenges of intensive memory costs and complex computation, making it a promising solution for on-device training, especially for resource-limited edge devices like FPGAs and ASICs.

However, it may be too early to draw this conclusion, as previous researchers have long overlooked a critical factor. That is, during the ZO optimization process, a different Gaussian random number needs to be added to each weight (to perturb the weight) in every inference. This implies that a large number of Gaussian random numbers are needed during every training step. This poses a significant challenge for hardware, as generating large quantities of random numbers, especially Gaussian random numbers, is not straightforward and can be very costly on embedded devices.
For example, implementing only one Gaussian random number generator (GRNG) on FPGA can cost 0.7\%\cite{thomas2015table} $\sim$ 10\%\cite{lee2006hardware} of FPGA resources (more details in Section~\ref{challenges}).
Having hundreds of GRNG on FPGA to generate random numbers is infeasible.
\textit{This critical hardware characteristic is often not recognized by algorithm designers, leading to a gap in compatibility between algorithm design and practical hardware implementation.}

\begin{table}[]
\centering
\vspace{27pt}
\caption{A comparison between existing approaches and the proposed method. MeZO: a representative ZO method for memory-efficient training but not taking on-device training scenarios into consideration.}
\scalebox{1}{
\begin{tabular}{lccc}
\toprule
Method               & \begin{tabular}[c]{@{}c@{}}Memory\\ efficiency\end{tabular} & \begin{tabular}[c]{@{}c@{}}Computational\\ simplicity\end{tabular} & \begin{tabular}[c]{@{}c@{}}Hardware\\ friendliness\end{tabular} \\ \hline
BP-based             &  \xmark                 &  \xmark     &  \xmark            \\
MeZO                 &  \cmark                 &  \cmark     &  \xmark            \\ \hline
\textbf{PeZO (ours)}          &  \cmark                 &  \cmark   &  \cmark              \\
\bottomrule
\end{tabular}
}
\label{compare}
\vspace{-25pt}
\end{table}


Is ZO optimization on hardware yet another failed idea that seems promising at first thought but actually not after deliberation?
To answer this question, this paper conducts a comprehensive study to explore whether feasible solutions exist to enable ZO optimization for on-device training on hardware. 
In this work, we use an FPGA as the proof-of-concept platform; however, the same issue also exists on other hardware devices, such as ASICs.
First, we revisited existing ZO optimization methods and conducted an in-depth discussion of the challenges associated with the hardware implementation.
We identified two major obstacles to hardware implementation: the need for a large quantity of random numbers and the requirement for costly Gaussian-distributed random numbers.
Then, to address these two obstacles, we propose our \textbf{P}erturbation-\textbf{E}fficient \textbf{Z}eroth-\textbf{O}rder (PeZO) optimization framework for hardware-friendly on-device training, the comparison of our proposal with existing approaches is shown in Table~\ref{compare}.
Specifically, we designed two random number reuse strategies tailored to two different random number generation settings, significantly reducing the number of random numbers required. 
Additionally, we designed a hardware-friendly adaptive modulus scaling method, which successfully enables the use of uniformly distributed random numbers as a substitute for costly Gaussian-distributed random numbers for ZO.
Note that naively replacing the Gaussian-distributed random numbers with uniform-distributed random numbers does not work and will lead to a severe accuracy drop (more details in Section~\ref{challenges}).

To the best of our knowledge, we are the first to explore the feasibility of implementing ZO on hardware, identify the critical yet severely overlooked issues, and propose effective solutions.
We have evaluated our proposed PeZO framework for fine-tuning real-world LMs and demonstrated that it achieves highly competitive results compared to the SOTA ZO method, MeZO~\cite{malladi2023fine}, which is infeasible for hardware implementation. 
In contrast, our PeZO successfully reduces the 48.6\% and 12.7\% of overall LUTs and FFs resource costs, and saves at a maximum of 86\% power consumption for random number generation, making ZO optimization feasible for on-device training.
We conducted comprehensive research on a series of critical issues for hardware implementation of ZO, including the impact of different random number reuse strategies on performance, the minimum number of random numbers required, and the impact of the number of bits in random numbers.
Moreover, we also summarize a comprehensive guideline and therefore provide valuable insights for future research in this area.

\section{Background and Challenges}

\subsection{Revisit Zeroth-order Optimization Fine-tuning}
\label{sec:revisit_ZO}


Recently, ZO optimization has gained significant attention in machine learning~\cite{verma2023certified,dhurandhar2019model,wang2022zarts,gu2021efficient}.
Unlike conventional first-order (FO) methods that compute gradients via backpropagation, ZO is gradient-free, estimating gradients using function value oracles and finite differences. This allows fine-tuning of deep neural networks (DNNs) with reduced memory costs~\cite{malladi2023fine}, requiring only two forward passes without storing memory-intensive activations or gradients.
MeZO~\cite{malladi2023fine} introduced a ZO-SGD algorithm for fine-tuning LMs, reducing memory usage by up to 12× without compromising accuracy. 
Table~\ref{ZOFOcompare} shows comparisons on memory costs and training FLOPs between the conventional backpropagation-based FO method and the ZO-based method on different sizes of OPT models.

The core idea of ZO optimization is to estimate gradients by applying random perturbations to the loss function and computing differences in function values. For a mini-batch of data $\mathcal{B}$, sampled from a labeled dataset $\mathcal{D} = {(x_{i}, y_{i})}_{i=1}^{|\mathcal{D}|}$, a model with parameters $\bm{\theta} \in \mathbb{R}^{d}$, where $d$ represents the dimension of the parameter space, and the corresponding loss function $\mathcal{L}(\bm{\theta}; \mathcal{B})$. The gradient is estimated as follows:
\begin{equation}
\label{equ:1}
\small
    \hat{\nabla} \mathcal{L}(\bm{\theta};\mathcal{B})=\frac{1}{q} \sum_{i=1}^q\left[\frac{\mathcal{L}\left(\bm{\theta}+\epsilon \hat{u}_{i};\mathcal{B}\right)-\mathcal{L}\left(\bm{\theta}-\epsilon \hat{u}_{i};\mathcal{B}\right)}{2 \epsilon} \hat{u}_{i}\right]
\end{equation}
where $\hat{u}_{i}$ is a random vector (\textbf{perturbation}) that has the same dimension as the model weights and is typically drawn from the standard Gaussian distribution $\mathcal{N}(0, \mathbf{I})$~\cite{malladi2023fine, zhang2024revisiting}, or Gaussian
sampling over a unit sphere with~\cite{liu2018zeroth, shamir2017optimal}, $q$ is the number of function queries, and $\epsilon > 0$ is a small smoothing parameter.

Given learning rate $\eta$ and mini-batch data $\mathcal{B}_{t}$ at training step $t$, the estimated gradient $\hat{\nabla} \mathcal{L} (\bm{\theta};\mathcal{B})$ can be obtained,
and ZO-SGD updates the parameters with the following rule:
\begin{equation}
\label{equ:2}
    \bm{\theta}_{t+1} = \bm{\theta}_{t} - \eta \hat{\nabla} \mathcal{L}(\bm{\theta};\mathcal{B}_{t})
\end{equation}

\begin{table}[t]
\centering
\renewcommand{\arraystretch}{0.9}
\vspace{13pt}
\caption{A memory and computation cost comparison between conventional backpropagation (BP)-based FO method and the ZO-based method for training different sizes of OPT models~\cite{zhang2022opt}. The FLOPs here indicate the per-iteration computation cost.
}
\begin{tabular}{lcccc}
\toprule
\multirow{2}{*}{\begin{tabular}[c]{@{}l@{}}Model\\ Size\end{tabular}} & \multicolumn{2}{c}{Memory cost} & \multicolumn{2}{c}{Train FLOPs} \\ \cline{2-5} 
                                                                          & BP-based         & ZO-based       & BP-based        & ZO-based       \\ \hline
1.3B                                                                      & 38.1 GB         & 2.6 GB          & 330.4 G         & 103.2 G            \\
2.7B                                                                      & 68.9 GB         & 5.4 GB           & 686.7 G         & 214.5 G           \\
6.7B                                                                      & 126.0 GB        & 13.4 GB           & 1756.6 G        & 549.8 G           \\
13B                                                                       & 213.0 GB        &   26.0 GB         & 3353.8 G          & 1048.6 G         \\ 
\bottomrule
\end{tabular}
\label{ZOFOcompare}
\vspace{-10pt}
\end{table}


The ZO optimization appears highly suitable for network fine-tuning on hardware, such as FPGAs and ASICs, as it not only reduces memory costs but also can fully reuse the inference processing elements (PEs) to estimate gradients without requiring reconfiguration. 
\textit{\textbf{However, the perturbation, a deceptively simple operation, could become the most significant challenge that prevents all existing ZO methods from being applied to hardware.}}


\subsection{Challenges of Applying ZO Optimization on Hardware}
\label{challenges}

As demonstrated in Equation (\ref{equ:1}), ZO optimization requires perturbing model $\bm{\theta}$ with a Gaussian random vector $\hat{u}_i$.
In other words, a different Gaussian random number needs to be added to each weight in every forward process. 
These challenges mainly stem from two requirements: (1) \textit{the necessity for Gaussian-distributed random numbers, and (2) \textit{the need for a large quantity of random numbers in every clock cycle}}.

Generating Gaussian-distributed random numbers on hardware is a very challenging task and usually requires the use of hardware-based random number generators (RNGs), which will inevitably introduce considerable hardware costs and 
complex operations.
The representative methods of Gaussian-distributed random number generators (GRNG) include 
Box-Muller~\cite{lee2006hardware}, 
Central Limited Theorem~\cite{thomas2014fpga}, 
TreeGRNG~\cite{crols2024treegrng},
T-Hadamard~\cite{thomas2015table}, 
etc.
For the precision-oriented GRNG design~\cite{lee2006hardware}, only one GRNG can take 3056 (6.6\%) FFs and 12 (10\%) DSPs on a Virtex-2 FPGA. 
Even for the efficiency-oriented GRNG design~\cite{thomas2015table}, it can still take 544 (0.7\%) FFs on a Virtex-6 FPGA.
Obviously, it is not feasible to have hundreds or thousands of GRNGs working in parallel to fulfill the computation parallelism (considering the tiling size of the 1024 or more to maximize the hardware-driven parallel matrix computation\cite{280896,10689464}).


Compared to the GRNG, a uniform-distributed random number generator (URNG) is relatively less complex. The linear-feedback shift register (LFSR) is a commonly used structure in URNG~\cite{na2017chip}, which takes several to tens of FFs depending on the bit-width of the generator~\cite{colavito2009efficient}.
If uniform-distributed random numbers can be used in place of Gaussian-distributed random numbers in ZO, it could alleviate the hardware overhead.
\textit{However, naively replacement
does not work, which will lead to a severe accuracy drop (refer Table~\ref{table1} and elaborated in Section~\ref{sec:RN_scaling}).}
Moreover, using a large number of URNGs to generate random numbers in parallel can still introduce non-negligible hardware costs~\cite{yuan2022you}.

\begin{table}[t]
\centering
\renewcommand{\arraystretch}{1.0}
\vspace{13pt}
\caption{Accuracy of ZO optimization using Gaussian-, uniform-distributed random number, and our proposed method (Section~\ref{sec:RN_scaling}). SST2 dataset and RoBERTa-large model are used. $k$ is the number of training samples per class. 
}
\label{table1}
\begin{tabular}{ccc}
\toprule
Perturbation distribution & $k=16$   & $k=256$  \\ \hline
Gaussian                  & 90.4 & 93.1 \\
Rademacher (+1, -1)                  & 51.9 & 50.5 \\
Uniform                   & 51.2 & 49.5 \\
Ours                      & 90.7     & 92.9    \\
\bottomrule
\end{tabular}
\vspace{-10pt}
\end{table}

\begin{figure*}[htbp]
\hspace{-1.3cm}
	\centering
	\begin{subfigure}{0.40\linewidth}
		\centering
		\includegraphics[width=1.3\linewidth]{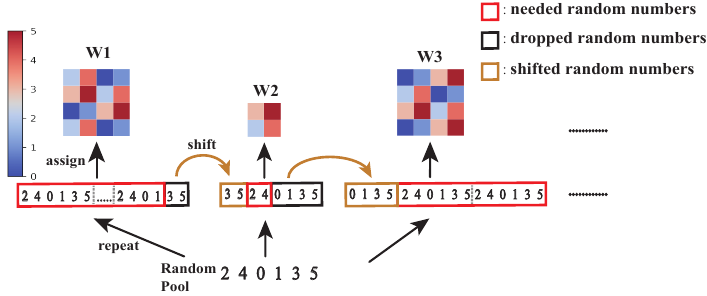}
		\caption{Pre-generation reuse strategy}
	\end{subfigure}
        \hspace{1.9cm}
	\begin{subfigure}{0.40\linewidth}
		\centering
		\includegraphics[width=1.2\linewidth]{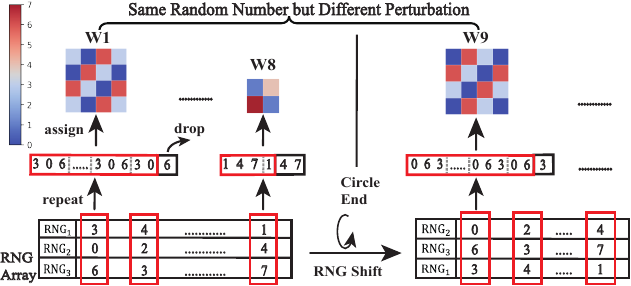}
		\caption{On-the-fly generation reuse strategy}
	\end{subfigure}
\vspace{10pt}
 \caption{Illustration of two random number reuse strategies for pre-generation and on-the-fly generation settings.}
 \label{Figure1}
 \vspace{5pt}
\end{figure*}

\subsection{Does CPU-based Generation Work?}
Given that many advanced FPGAs are now equipped with CPUs to handle complex applications, and CPUs are generally considered outperform in generating random numbers than traditional FPGAs, one might question whether we can simply use the CPU to quickly generate a large number of random numbers to meet the requirements of ZO optimization.

Unfortunately, our latency tests indicate that CPU-based random number generation is still insufficient for ZO optimization. Specifically, for the well-known LM LLaMA2-7B, updating a single attention layer requires 4$\times$4096$\times$4096 random numbers for perturbation. 
On an Arm Cortex-A53 CPU embedded inside the AMD Xilinx ZCU102 FPGA development platform, generating Gaussian random numbers for this purpose takes 11,927.258 ms. 
In contrast, FPGA-based inference time for such a layer takes only 2.013 ms at a frequency of 225 MHz~\cite{zeng2024flightllm}\footnote{Reported inference time is scaled in consideration of the hardware resource difference between Xilinx ZCU102 and Alveo U280 in the original work.}.
This results in a minimum latency performance margin of 5900×, making real-time CPU-based random number generation an unacceptable solution for ZO fine-tuning—even before considering the additional transfer latency of over 40 ms from the AXI interconnect~\cite{9218652}. 

\textbf{Faced with the above dilemma, we pose the following question: \emph{Can we use fewer random numbers and also avoid the costly Gaussian distribution?} The answer is affirmative with our proposed PeZO framework.}
\section{Design of Perturbation-efficient ZO Optimization Framework}
\label{pezo}


To address the aforementioned question, an idea is whether, instead of assigning a unique random number to each weight, we could allow different weights to share and reuse the same random number appropriately. This would reduce the number of distinct random numbers required at any clock cycle.
Although the idea appears straightforward, it involves addressing several complex challenges, for instance, (1) determining how to efficiently reuse the random numbers, (2) how many random numbers are needed, and (3) how many bits are enough.
To answer these questions, we first explore two random number generation settings and design two corresponding reuse strategies to minimize the required number of random numbers while ensuring the irregularity and diversity of the perturbations. (Section~\ref{sec:RN_reuse}).
Then, we designed a hardware-friendly adaptive modulus scaling method, enabling uniform distribution-based perturbation in ZO optimization to replace the costly Gaussian distribution-based perturbation (Section~\ref{sec:RN_scaling}).
Putting it all together, we answer all the aforementioned questions and propose our  \textbf{P}erturbation-\textbf{E}fficient \textbf{Z}eroth-\textbf{O}rder (PeZO) optimization framework for hardware-friendly on-device training.



\subsection{Design of Two Random Number Reuse Strategies}
\label{sec:RN_reuse}

The previous works~\cite{aghajanyan2020intrinsic} find that LMs have low intrinsic dimensions, which means the optimal parameter adjustments lie on low-dimensional manifolds within the high-dimensional space. This provides us with the potential opportunity to optimize with less diverse perturbation (fewer random numbers).

\noindent\textbf{Two generation settings.}
We explore two random number generation settings for ZO optimization: \textit{pre-generation} and \textit{on-the-fly generation}. 
\underline{For the pre-generation}, 
random numbers are generated in advance and pre-stored on FPGA. These pre-generated numbers are reused throughout training without introducing new ones. The setting simplifies design complexity, eliminates the need for hardware-based RNGs, and reduces continuous generation costs.
\underline{For the on-the-fly generation}, it involves RNGs to dynamically generate random numbers during training, eliminating the need for storage while enhancing irregularity and diversity of perturbations.
For each generation setting, we design a tailored reuse strategy, termed pre-generation-based reuse and on-the-fly-based reuse.
 
\noindent\textbf{Two tailored reuse strategies.}
Our goal in designing the reuse strategies is to maximize the diversity and irregularity of perturbations while minimizing the use of unique random numbers, all without harming hardware-friendliness. This is crucial as the diversity and irregularity of perturbations significantly affect the performance of ZO optimization. The design of two reuse strategies is shown in Figure~\ref{Figure1}. For simplicity, we denote these two reuse strategies as pre-generation strategy and on-the-fly generation strategy.
\underline{For the pre-generation}, $N$ random numbers sampled from~$\mathcal{U}(-1, 1)$ are pre-generated to create a pool. Perturbations are created by repeatedly concatenating numbers from this pool to the required length.
\underline{For the on-the-fly generation}, $n$ RNGs produce $n$ distinct random numbers per clock cycle, which are concatenated similarly to the pre-generation method. On FPGA, buffers or shift registers store and assemble these numbers efficiently, ensuring low latency and minimal resource usage.


Notably, we avoid setting the random number pool size or the number of RNGs to exact powers of two to prevent regular patterns in perturbation, as the number of weights in each layer is also typically powers of two.
Such regularity could undermine the irregularity and diversity required for effective ZO optimization. Therefore, random numbers or RNGs are set to powers of two minus one (i.e., $2^{n}-1$), but denote them as powers of two for simplicity in the following sections. 

To further enhance the irregularity and diversity of generated perturbations, we implement different shift operators on the random number pool and RNGs. \underline{For the pre-generation}, as the number of model weights $|\bm{\theta}|$ is not divisible by the pool size, leftover random numbers remain after each perturbation. We shift these leftovers to the beginning of the next perturbation and fill the end with numbers from the random number pool.
\underline{For the on-the-fly generation}, each RNG produces one random number at a time, yielding $2^b$ unique random numbers per cycle, where $b$ is the bit-width of RNGs. Consequently, the RNGs array generates $2^b$ possible different random number combinations. To increase irregularity and diversity, we shift the RNG that generates the first-position random number to the end of the RNG array after each clock cycle. This increases the combination number of random numbers to $n\times2^b$, where $n$ is the number of RNGs. The shifting mechanism can be efficiently implemented using circular buffers or shift registers, leveraging BRAM and parallel processing to minimize resource usage and latency.

The perturbation in a ZO optimization step is linearly related to the direction and magnitude of the weights update (Equation~(\ref{equ:1}) and (\ref{equ:2})). Heuristically, this means that the pre-generation method will explore a larger loss landscape because it contains more unique random numbers in perturbations than the on-the-fly generation method (i.e., $n\ll N$). These differing characteristics of reuse strategies result in different optimization patterns, leading to varying preferences depending on the task. We will analyze these preferences in the strategy comparison section (in Section~\ref{sec:closer_look}).

\subsection{Design of Hardware-friendly Adaptive Modulus Scaling}
\label{sec:RN_scaling}


As discussed in Section~\ref{challenges} and Table~\ref{table1}, directly replacing Gaussian-distributed random numbers with uniform-distributed random numbers will lead to a severe accuracy drop. 
One main reason is that the large integers in originally generated uniform random numbers can lead to an overly significant perturbation, collapsing the model training.

A potential solution is to scale the random numbers based on their precise modulus value. However, since the modulus varies with each set of generated random numbers and is not fixed, computing it on the fly is hardware-unfriendly (as elaborated later). 
Alternatively, using a fixed scaling factor calculated from the expected statistical modulus over multiple rounds of generated random numbers may reduce computational complexity but introduce substantial scaling errors.
These errors are particularly severe when using a smaller number of random numbers.
Therefore, we propose an adaptive scaling method that minimizes the modulus scaling error while maintaining hardware-friendliness.

Specifically, we propose scaling the modulus of generated uniform random numbers to match the expected modulus of a Gaussian perturbation with the same dimension.
In this way, the proper perturbations can be maintained to provide necessary coverage of the manifold while avoiding model collapse~\cite{malladi2023fine, zhang2024revisiting}.
Given the generated uniform perturbation $u_{i}$, where $i$ indicates $i$-th query for loss function. Let its modulus match the same-dimensioned Gaussian perturbation:
\begin{equation}
\bar{u}_{i} = \frac{\mathbb{E}\|\hat{u}_{i}\|_2}{\|u_{i}\|_2} \times u_{i}
\end{equation}
where $\bar{u}_{i}$ is the scaled uniform perturbation, $\hat{u}_{i}$ is a Gaussian perturbation with the same size of $\bar{u}_{i}$, and each component of $\hat{u}_{i}$ is drawn from a standard Gaussian distribution, i.e., $ \hat{u}_{ij} \in \mathcal{N}(0,1)~\text{for}~j=1,2,\dots,d$, with $d$ being the dimension. 

The modulus expectation of a Gaussian perturbation of dimension $d$ is computed as:
\begin{equation}
\mathbb{E}\|\hat{u}_{i}\|_2 = \sqrt{2} \frac{\Gamma((d+1) / 2)}{\Gamma(d / 2)}
\end{equation}
where $\Gamma(\cdot)$ denotes the Gamma function~\cite{wishart1928generalised}.
To prevent the computation overflow caused by a large value of $d$ in the Gamma function, 
we reformulate the expected modulus as:
\begin{equation}
\footnotesize
\mathbb{E}\|\hat{u}_{i}\|_2 = \exp{\left( 0.5 \cdot \log(2) + \log\Gamma\left(\frac{d + 1}{2}\right) - \log\Gamma\left(\frac{d}{2}\right) \right)}
\end{equation}


Another key challenge to scale the perturbation is to adaptively compute the scaling factor $\frac{\mathbb{E}\|\hat{u}_{i}\|_2}{\|u_{i}\|_2}$. Due to the involvement of FPGA-unfriendly operations like division, $\log(\cdot)$, $\exp(\cdot)$, and the gamma function $\Gamma(\cdot)$. Implementing these operations on hardware would consume significant computational resources. 
For the pre-generation method, we can scale the random numbers in advance before storing them on hardware.
So, we mainly focus on addressing the dynamic scaling in the on-the-fly generation method.   
We devise computational tricks for both $\mathbb{E}\|\hat{u}_{i}\|_2$ and $\|{u}_{i}\|_2$ respectively as follows.

For $\mathbb{E}\|\hat{u}_{i}\|_2$, it is a fixed statistical value with respect to the number of RNGs used. 
We compute this value before training and then store it in the memory, avoiding repeated computation and saving resources.

\begin{figure}[!t]
 \centering
 \hspace{-0.4cm}
 \includegraphics[width=0.9\linewidth]{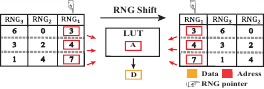}
 \caption{Illustration of using the random number generated by the RNG pointed to by the pointer to efficiently query the scaling factor from the look-up table.
 }
 \label{mapping}
\end{figure}

For $\|{u}_{i}\|_2$, as each RNG generates one random number per clock cycle, resulting in $2^{b}$ possible combinations (length of one cycle), where $b$ is the bit width of the RNGs. Though shift operations (in our reuse strategy) increase the number of combinations, they do not affect the modulus. Thus, we pre-compute the scaling factor $s_{i} = \frac{\mathbb{E}\|\hat{u}_{i}\|_2}{\|u_{i}\|_2}$ of different combinations (i.e., different ${u}_{i}$) and store them in a look-up table (LUT) of size $2^{b}$, implemented in the FPGA's block RAM for efficient storage and fast access. 
An RNG pointer tracks the position of the first RNG, and its current output serves as the address to query the LUT for the scalar. Figure~\ref{mapping} illustrates this process of query. Using the FPGA's parallel processing capabilities, the LUT is accessed concurrently with other operations to minimize latency. In addition, we round the scale factor to the nearest power of two before storing so that the scaling process only needs to conduct bitwise operations. 

\begin{table*}[ht]
\centering
\caption{Overall experiment results on RoBERTa-base (125M parameters) and RoBERTa-large (350M). All reported numbers are averaged accuracy with standard deviation shown. Two reuse strategies both show competitive performance compared with baseline training, but with much less random number requirement. Better results among ZO-based methods are highlighted in bold.}
\label{table3}
\renewcommand{\arraystretch}{1}
\scalebox{0.95}{
\begin{tabular}{llccccc}
\toprule
\multirow{2}{*}{Configuration}                                                   & \multirow{2}{*}{\begin{tabular}[c]{@{}l@{}}Task \\ Type\end{tabular}} & \textbf{SST-2}      & \textbf{SST-5}      & \textbf{MNLI}          & \textbf{RTE}          & \textbf{TREC}                   \\
                                                                                 &                                                                       & \multicolumn{2}{c}{------sentiment------} & \multicolumn{2}{c}{----language inference----} & \multicolumn{1}{c}{---topic---} \\ \hline
\multirow{4}{*}{\begin{tabular}[c]{@{}l@{}}$k=16$ \\ RoBERTa-base\end{tabular}}   & BP-based                                                                    & 88.1 (2.2)                           & 40.3 (1.7)                           & \textbf{55.7} (2.4) & 56.6 (4.3)                           & \textbf{59.6} (3.1) \\
                                                                                  & MeZO~\cite{malladi2023fine}                                                                  & 88.1 (2.2)                           & 40.3 (1.7)                           & \textbf{55.7} (2.4) & 56.6 (4.3)                           & \textbf{59.6} (3.1) \\
& Ours Pre-generation                                                        & 87.8 (2.2)          & 42.1 (1.8)          & 53.2 (2.5)             & 54.5 (4.7)            & 58.2 (2.6)                      \\
                                                                                 & Ours On-the-fly generation                                                 & \textbf{88.4} (2.8)          & \textbf{42.4} (1.8)          & 54.8 (2.9)             & \textbf{57.3} (4.0)            & 58.6 (3.8)                      \\ \hline
\multirow{4}{*}{\begin{tabular}[c]{@{}l@{}}$k=256$ \\ RoBERTa-base\end{tabular}}   & BP-based                                                                    & 88.1 (2.2)                           & 40.3 (1.7)                           & \textbf{55.7} (2.4) & 56.6 (4.3)                           & \textbf{59.6} (3.1) \\
                                                                                  & MeZO~\cite{malladi2023fine}                                                                   & 89.3 (1.4)          & 47.3 (1.2)          & 58.3 (1.8)             & 65.7 (2.1)            & 81.4 (0.9) \\
                                                                                 & Ours Pre-generation                                                        & \textbf{89.4} (1.0)          & \textbf{47.6} (0.9)          & 58.1 (1.3)             & \textbf{71.5} (2.4)            & \textbf{81.9} (1.1)                      \\
                                                                                 & On-the-fly generation                                                 & 88.2 (1.9)          & 44.8 (1.3)          & \textbf{59.7} (2.3)             & 60.3 (1.7)            & 78.4 (1.2)                      \\ \hline \hline
\multirow{4}{*}{\begin{tabular}[c]{@{}l@{}}$k=16$ \\ RoBERTa-large\end{tabular}}   & BP-based                                                                    & 91.9 (1.8)                           & 47.5 (1.9)                           & 70.0 (2.3) & 66.4 (7.2)                           & 85.0 (2.5) \\
                                                                                  & MeZO~\cite{malladi2023fine}                                                                  & 90.4 (1.3)          & \textbf{45.3} (1.8)          & 59.3 (2.3)             & 59.9 (4.3)            & 73.0 (4.2) \\
                                                                                 & Ours Pre-generation                                                        & 90.7 (1.2)          & 44.7 (2.0)          & 60.1 (2.5)             & 63.2 (4.9)            & 72.0 (3.2)                      \\
                                                                                 & Ours On-the-fly generation                                                 & \textbf{90.9} (1.4)          & 44.8 (1.9)          & \textbf{61.1} (2.4)             & \textbf{63.9} (4.4)            & \textbf{76.2} (2.4)                      \\ \hline
\multirow{4}{*}{\begin{tabular}[c]{@{}l@{}}$k=256$ \\ RoBERTa-large\end{tabular}}   & BP-based                                                                    & 93.9 (0.7)                           & 55.9 (0.9)                           & 84.4 (0.8) & 82.7 (1.4)                           & 97.3 (0.2) \\
                                                                                  & MeZO~\cite{malladi2023fine}                                                                  &\textbf{93.1} (0.5)          & \textbf{50.0} (0.9)          & 72.8 (0.7)             & 72.2 (1.5)            & \textbf{90.8} (0.9) \\
                                                                                 & Ours Pre-generation                                                        & 92.9 (0.6)          & 49.4 (0.8)          & \textbf{73.6} (1.3)             & \textbf{72.5} (1.1)            & 89.8 (1.1)                      \\
                                                                                 & Ours On-the-fly generation                                                 & 91.9 (1.1)          & 48.3 (1.7)          & 68.7 (1.2)             & 66.4 (2.4)            & 84.2 (1.6)  \\
\bottomrule                                                                                
\end{tabular}
} 
\end{table*}

\begin{table*}[ht]
\centering
\vspace{10pt}
\caption{Overall experiment results on fine-tuning OPT and Llama series models. Two reuse strategies both show competitive performance compared with baseline training, but with much less random number requirement. Better results among these ZO methods are highlighted in bold.
}
\label{table4}
\renewcommand{\arraystretch}{1}
\scalebox{0.9}{
\begin{tabular}{llccccc}
\toprule
\multirow{2}{*}{\begin{tabular}[c]{@{}l@{}}Model \\ Type\end{tabular}} & \multirow{2}{*}{\begin{tabular}[c]{@{}l@{}}Task\\ Type\end{tabular}} & \textbf{SST-2}    & \textbf{RTE}    & \textbf{WIC}    & \textbf{WSC}   & \textbf{COPA}           \\
                                                                       &                                                                      & \multicolumn{4}{c}{------------classification------------} & -multiple choice- \\ \hline
\multirow{4}{*}{OPT-1.3B}                                              
& BP-based                                                                 & 90.6              & 59.3            & 59.7            & 57.0           & 84.0                    \\
                                                                       & MeZO~\cite{malladi2023fine}                                                                 & 85.6              & 55.9            & \textbf{60.3}            & 53.8           & 76.0                    \\
                                                                       & Ours Pre-generation                                                       & 86.4              & 53.3            & 59.7            & 53.9           & 75.0                    \\
                                                                       & Ours On-the-fly generation                                                & \textbf{87.2}              & \textbf{57.0}            & 59.4            & \textbf{56.7}           & \textbf{78.0}                    \\ \hline
\multirow{4}{*}{OPT-2.7B}                                             
& BP-based                                                                 & 93.6              & 63.5            & 58.8            & 59.3           & 80.0                    \\
                                                                       & MeZO~\cite{malladi2023fine}                                                                 & \textbf{92.2}              & 53.1            & 57.4            & 47.1           & 75.0                    \\
                                                                       & Ours Pre-generation                                                       & 91.4              & 52.4            & \textbf{61.6}            & 50.0           & 76.0                    \\
                                                                       & Ours On-the-fly generation                                                & 91.9              & \textbf{56.3}            & 60.3            & \textbf{57.9}           & \textbf{76.0}      \\ \hline \hline
\multirow{4}{*}{Llama3-1B}
& BP-based                                                                 & 89.8              & 60.6            & 62.3            & 58.8           & 84.0                    \\
                                                                       & MeZO~\cite{malladi2023fine}                                                                 & 86.9              & 57.1            & 60.4            & 52.9           & 73.0                    \\
                                                                       & Ours Pre-generation                                                       & \textbf{88.0}              & 57.4            & 60.1            & 53.8           & \textbf{78.0}                    \\
                                                                       & Ours On-the-fly generation                                                & 87.5              & \textbf{58.3}            & \textbf{61.3}            & \textbf{55.6}           & 76.0      \\ \hline
\multirow{4}{*}{Llama3-3B}
& BP-based                                                                 & 92.9              & 67.4            & 63.2            & 59.8           & 87.0                    \\
                                                                       & MeZO~\cite{malladi2023fine}                                                                 & 90.0              & \textbf{59.1}            & 59.2            & 54.8           & 75.0                    \\
                                                                       & Ours Pre-generation                                                       & \textbf{90.7}              & 58.9            & \textbf{60.1}            & 56.6           & \textbf{76.0}                    \\
                                                                       & Ours On-the-fly generation                                                & 89.5              & 58.8            & 57.8            & \textbf{58.4}           & 73.0      \\
\bottomrule
\end{tabular}
}
\end{table*}

\begin{figure*}[htbp]
	\centering
        \begin{minipage}{0.32\linewidth}
		\centering
		\includegraphics[width=1\linewidth]{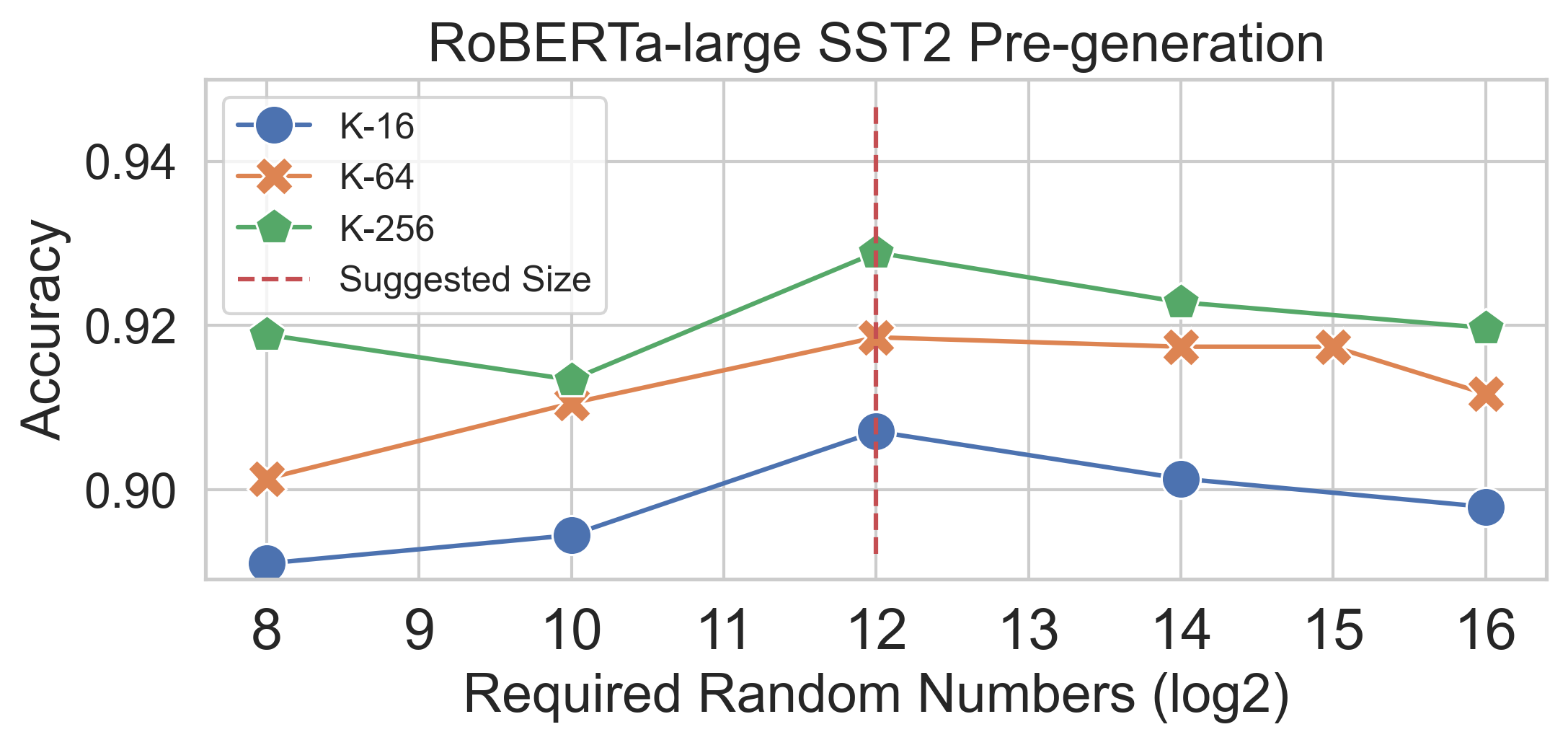}
	\end{minipage}
        \begin{minipage}{0.32\linewidth}
		\centering
		\includegraphics[width=1\linewidth]{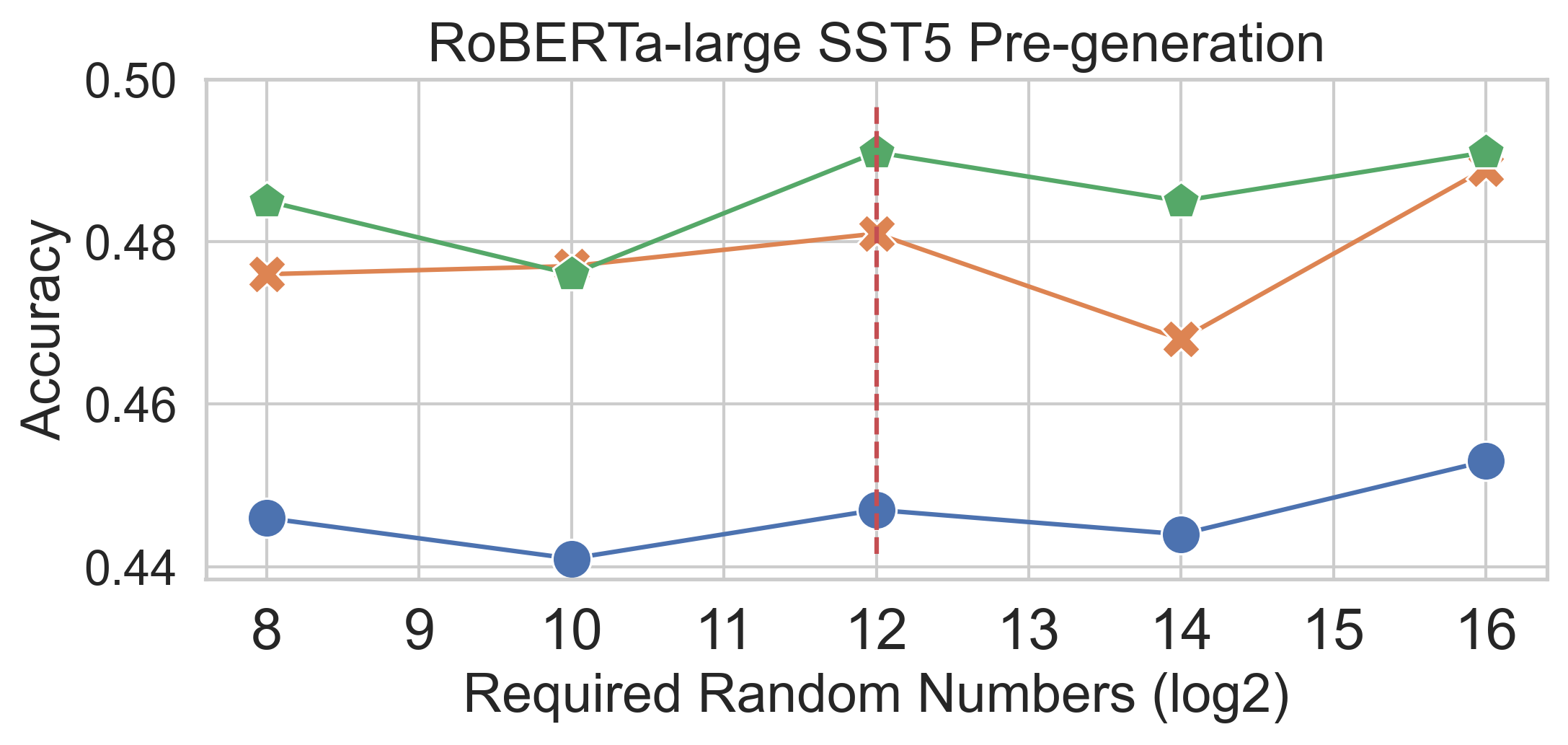}
	\end{minipage}
	\begin{minipage}{0.32\linewidth}
		\centering
		\includegraphics[width=1\linewidth]{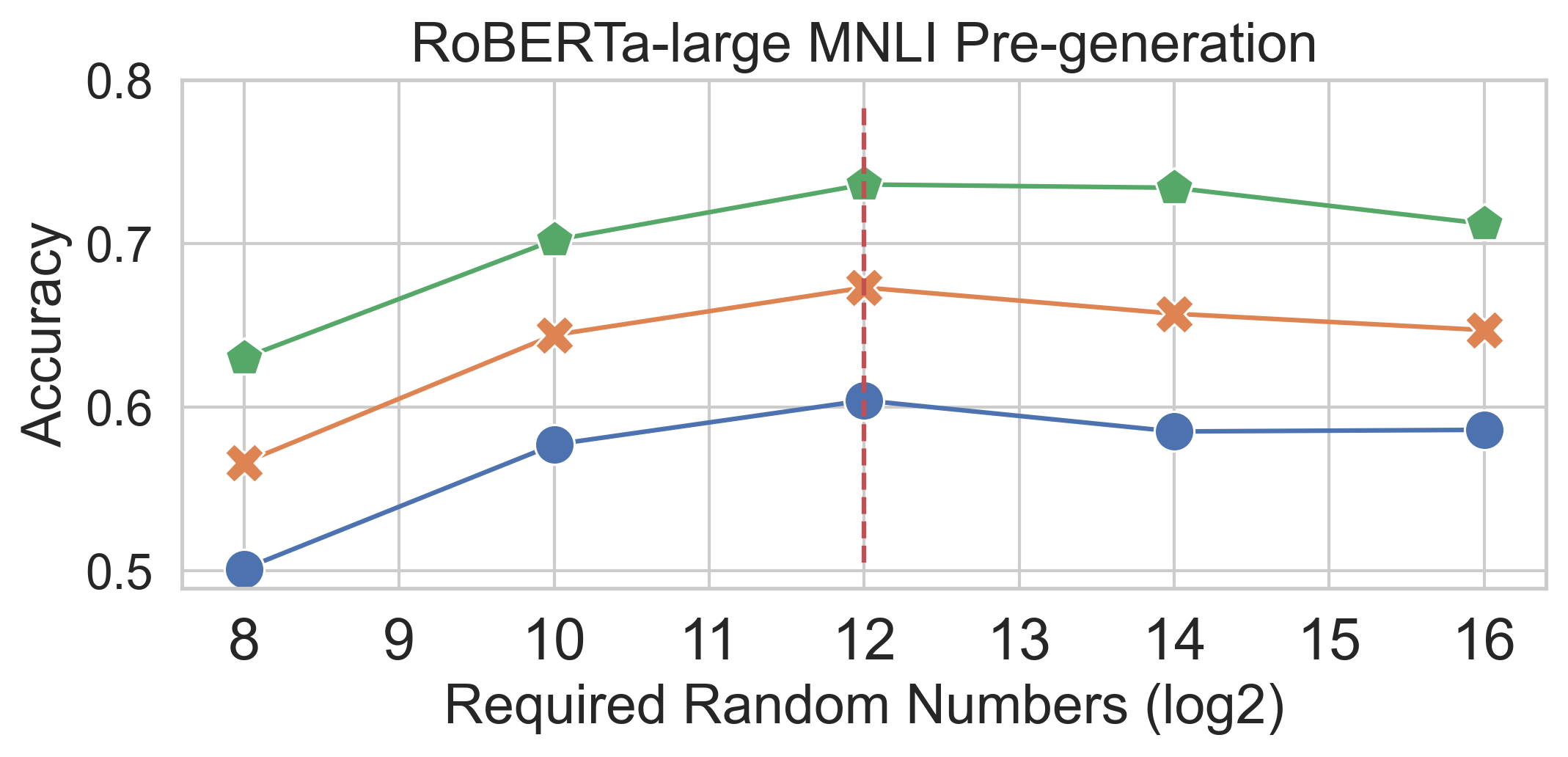}
	\end{minipage}

	\qquad
 
	\begin{minipage}{0.32\linewidth}
		\centering
		\includegraphics[width=1\linewidth]{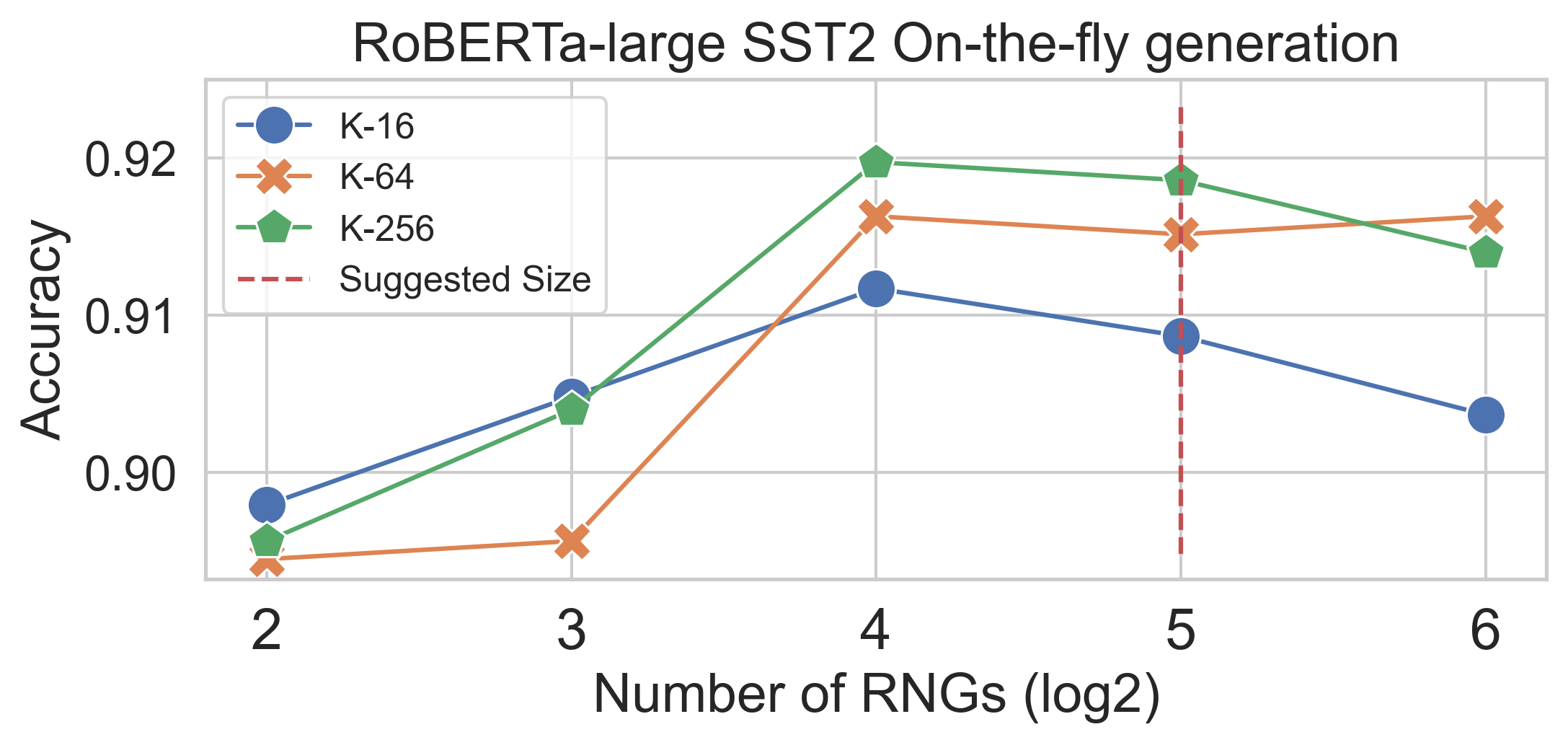}
	\end{minipage}
	\begin{minipage}{0.32\linewidth}
		\centering
		\includegraphics[width=1\linewidth]{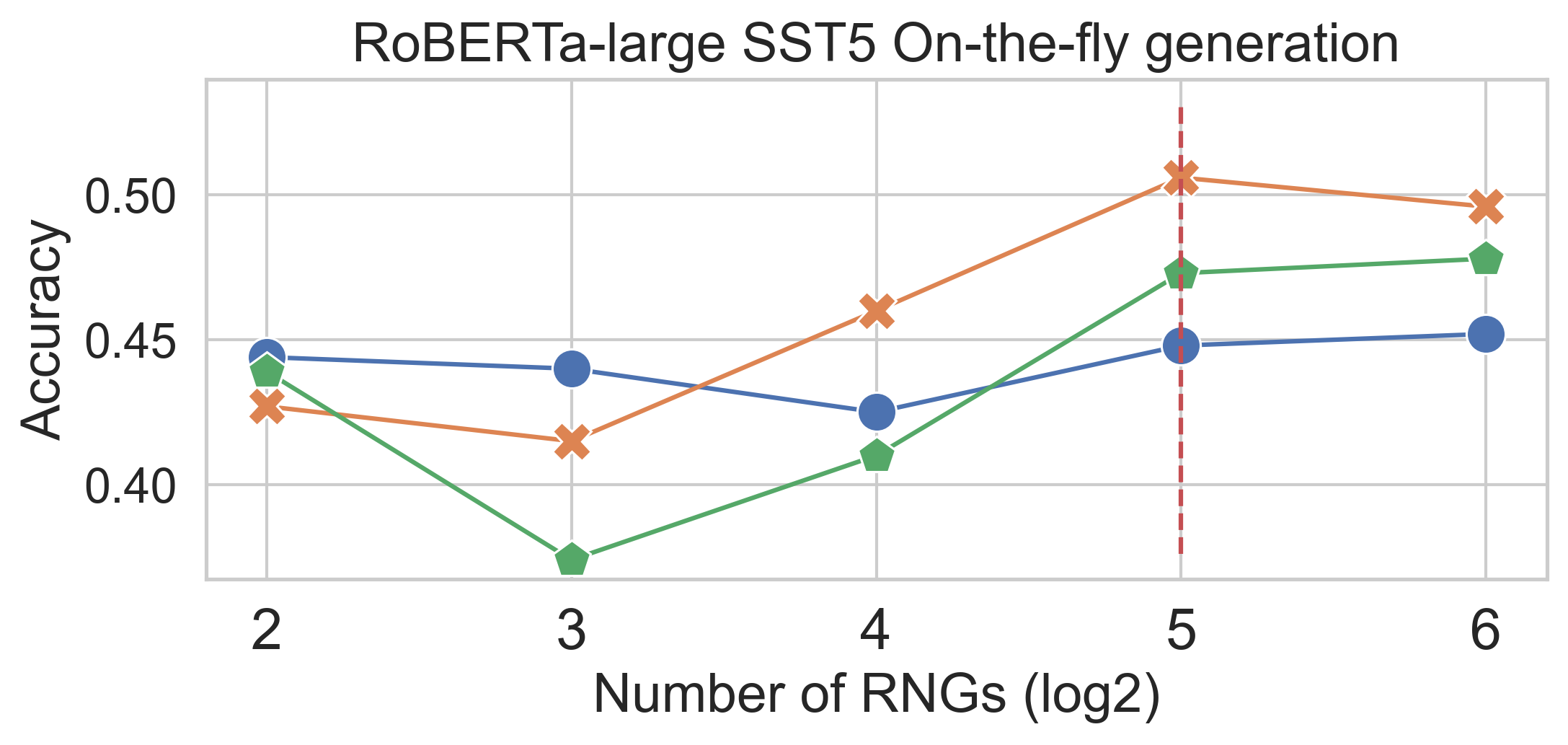}
	\end{minipage}
        \begin{minipage}{0.32\linewidth}
		\centering
		\includegraphics[width=1\linewidth]{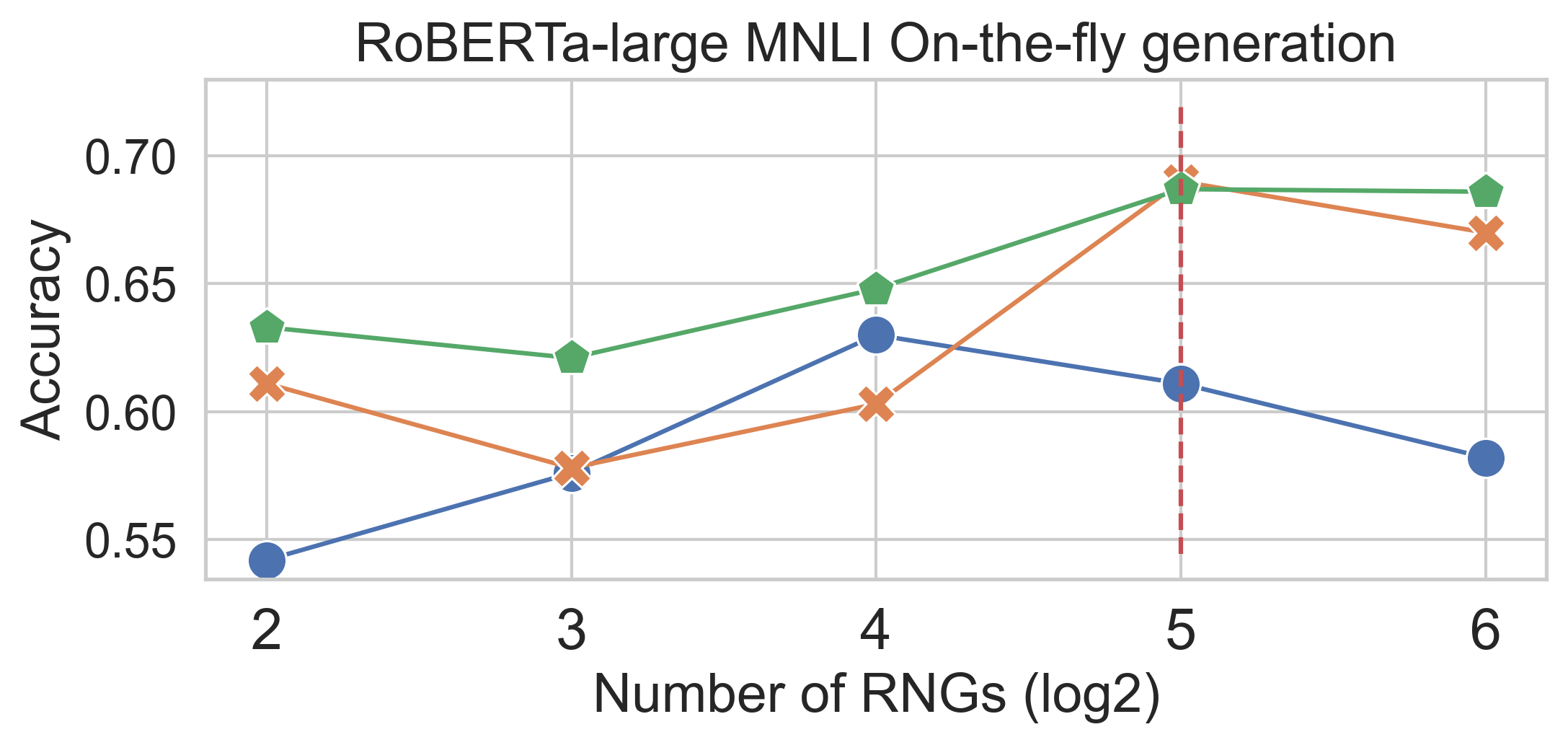}
	\end{minipage}

        \qquad

        \begin{minipage}{0.32\linewidth}
		\centering
		\includegraphics[width=1\linewidth]{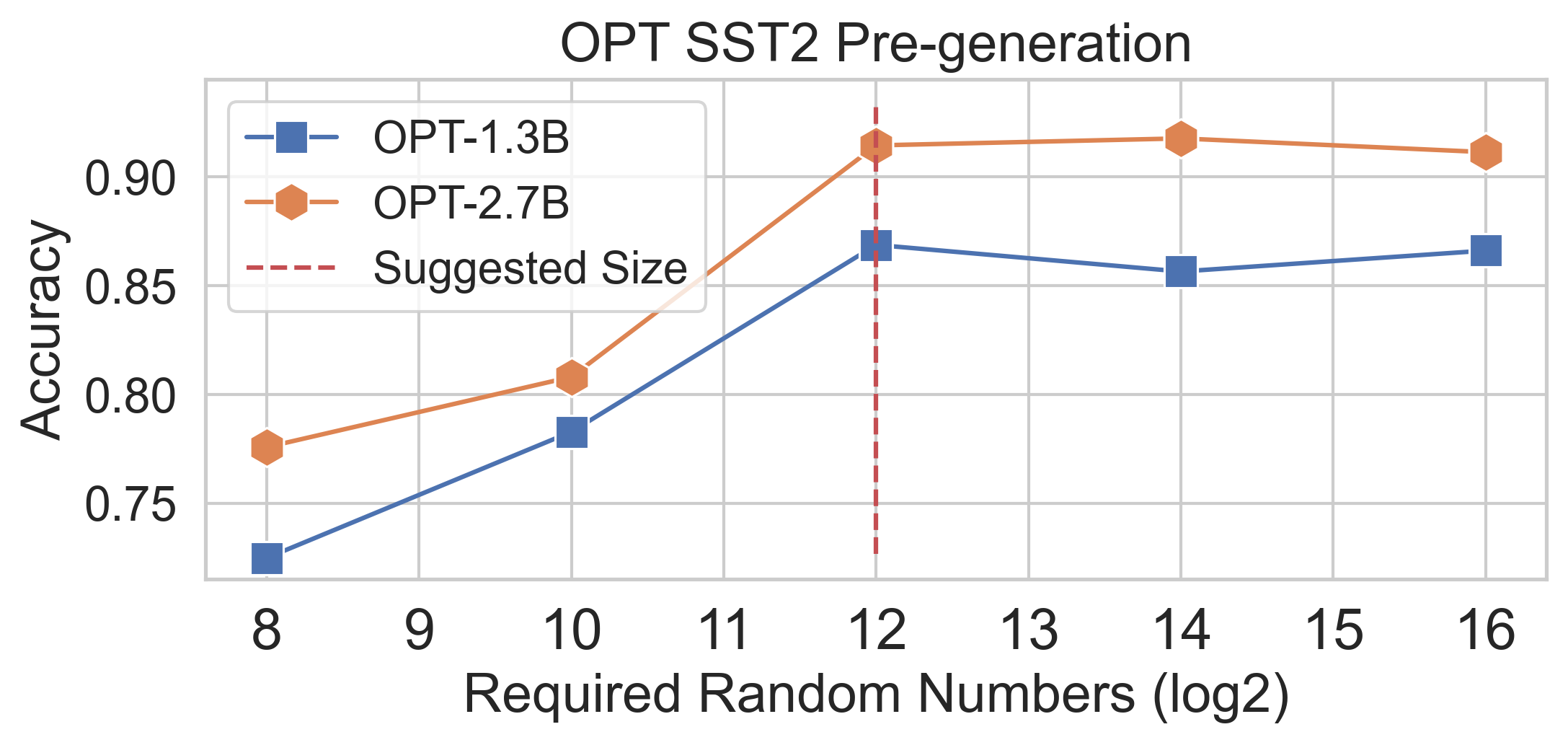}
	\end{minipage}
	\begin{minipage}{0.32\linewidth}
		\centering
		\includegraphics[width=1\linewidth]{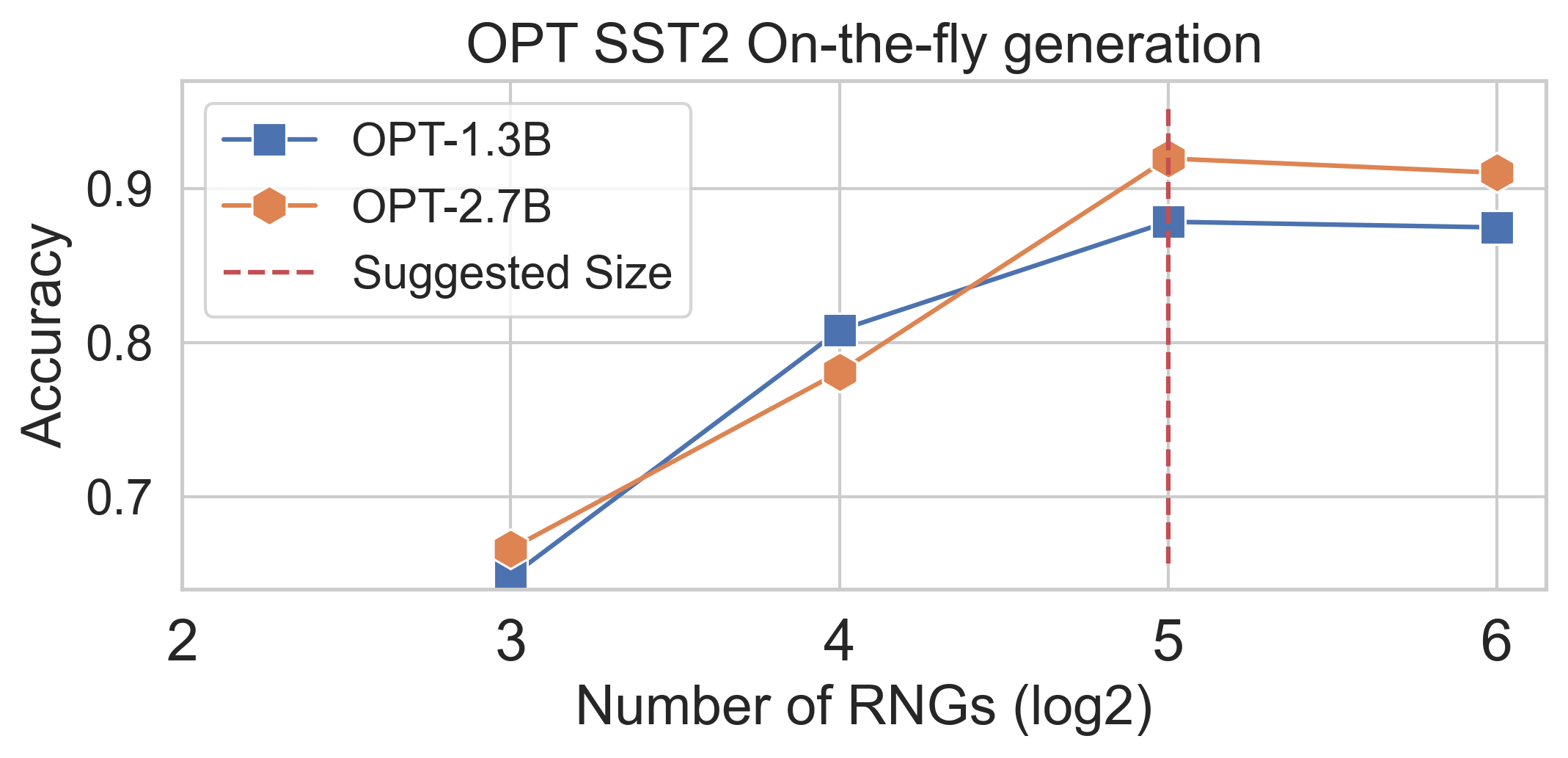}
	\end{minipage}

 \caption{Experiments of pre-generation strategy and on-the-fly generation strategy with different numbers of pre-generated random numbers/RNGs. OPT-series models collapse on SST2 with $2^{2}$ RNGs, thus there is no data point.}
 \label{Figure2}
\vspace{-5pt}
\end{figure*}
\section{Evaluation, Discussion, and Guideline}

Based on the proposed PeZO framework, we conduct comprehensive experiments with different LMs, datasets, and hyperparameter settings. Detailed analysis is demonstrated for a deeper understanding of conducting hardware-friendly and memory-efficient on-device training via zeroth-order optimization. 

\subsection{Experimental Settings}

\textbf{Models and datasets.} We evaluate PeZO with various LMs, including medium-sized masked models~\cite{liu2019roberta} (RoBERTa-base, RoBERTa-large) and large-sized autoregressive models~\cite{zhang2022opt} (OPT-1.3B, OPT-2.7B), total parameters size ranging from 125 million to 2.7 billion. 
For RoBERTa series, we evaluate with datasets: SST-2~\cite{socher2013recursive}, SST-5~\cite{socher2013recursive}, TREC~\cite{voorhees2000building}, MNLI~\cite{yao2020pyhessian}, and RTE~\cite{dagan2005pascal,bar2006second,bentivogli2009fifth,giampiccolo2007third}. 
For OPT series, SST2~\cite{clark2019boolq}, RTE~\cite{dagan2005pascal,bar2006second,bentivogli2009fifth,giampiccolo2007third}, WIC~\cite{pilehvar2018wic}, WSC~\cite{levesque2012winograd}, and COPA~\cite{roemmele2011choice} are evaluated.

\textbf{Baseline.} Our primary baseline is MeZO~\cite{malladi2023fine}, which is one of the state-of-the-art zeroth-order LMs fine-tuning methods but each weight needs a different random number for perturbation. The performance of the method represents an ideal perturbation condition in ZO optimization.

\textbf{Evaluation.} For training and evaluation, we follow previous work in studying the few-shot settings \cite{malladi2023fine,gao2020making}, randomly sampling $k$ samples per class for training and validation, and sampling 1000 samples for test. For the RoBERTa series model, we evaluate $k=16$ and $k=256$ settings. For the OPT series, we only evaluate the $k=16$ setting, considering real-world hardware platforms may not be able to afford that large computational overhead when setting a larger $k$. 
The Verilog-written RTL-based hardware design for this work is simulated and synthesized on AMD Xilinx Vivado 2020.1, and finally implemented on Xilinx ZCU102 FPGA development platform.  The Switching Activity Interchange Format (SAIF)-based measurement method is adopted inside Vivado to capture the fine-grained switch activity during the runtime for more accurate power consumption analysis.


\subsection{Closer Look at PeZO}
\label{sec:closer_look}

\noindent\textbf{Overall results.} We present our overall results in Table~\ref{table3} and Table~\ref{table4}. When comparing our framework with the baseline method MeZO~\cite{malladi2023fine}, we observe that both reuse strategies achieve comparable accuracy to the baseline. Specifically, PeZO attains a performance close to that of the baseline when using $k=16$. Furthermore, with $k=256$, the accuracy gap between PeZO and the baseline still narrows to within 0.5\% across all tasks. 


\noindent\textbf{Which reuse strategy is better?}

In terms of comparing two reuse strategies in accuracy, on-the-fly generation performs better in easier optimization scenarios (e.g., with smaller training data like $k=16$), while pre-generation excels in more challenging cases (e.g., with larger training data like $k=256$). This pattern holds consistently across datasets and models. For small datasets, the optimal solution is more likely to reside on a low-dimensional manifold~\cite{sener2020learning}, aligning with on-the-fly generation, which operates on fewer unique random numbers and optimizes on a lower-dimensional surface, leading to better results. In contrast, as more unique random numbers are provided, the pre-generation method tends to apply more diverse searching directions in one iteration and therefore performs better in the more challenging optimization settings. The hardware resources comparison between these two reuse strategies will be shown latter.




\noindent\textbf{How many random numbers or RNGs are needed at least?}

We investigate the number of random numbers or RNGs required to achieve optimal performance in pre-generation and on-the-fly generation. Specifically, we aim to determine how many weights can share the same random number without sacrificing the performance of the two reuse strategies.
For the pre-generation method, we use different sizes of random number pools ranging from $2^{8}$ to $2^{16}$. For the on-the-fly generation method, we vary the number of RNGs from $2^{2}$ to $2^{6}$. We conduct experiments on various datasets, training sample sizes, and language models.

The results in Figure~\ref{Figure2} show that for the pre-generation method, performance improves as the random number pool size increases from $2^{8}$ to $2^{12}$. However, expanding the pool beyond $2^{12}$ yields no significant additional performance gains. Interestingly, this pattern is consistent across all settings, as $2^{12}$ random numbers also suffice for fine-tuning larger LMs or datasets. 
Storing $2^{12}$ pre-generated random numbers only requires modest on-chip memory (BRAM), making it feasible for modern FPGAs.
For the on-the-fly generation method, peak performance occurs when using $2^{5}$ RNGs, and this trend remains consistent across settings. Additionally, for some tasks and datasets, the accuracy variance is pretty small given different random number settings, therefore, there is room for accuracy and hardware trade-off for flexible design implementation.


Our work reveals that the random numbers can be reused and shared among different weights for ZO fine-tuning.
Even with an extremely limited number of random numbers or RNGs (i.e., $2^{8}$ random numbers or $2^{2}$ RNGs), LMs can still be trained to an acceptable performance level. Based on these findings, we adopt $2^{12}$ random numbers and $2^{5}$ RNGs as the default settings for the pre-generation and on-the-fly generation methods in the subsequent experiments.

\begin{figure}[t]
\vspace{15pt}
	\centering
        \begin{minipage}{0.49\linewidth}
		\centering
		\includegraphics[width=1\linewidth]{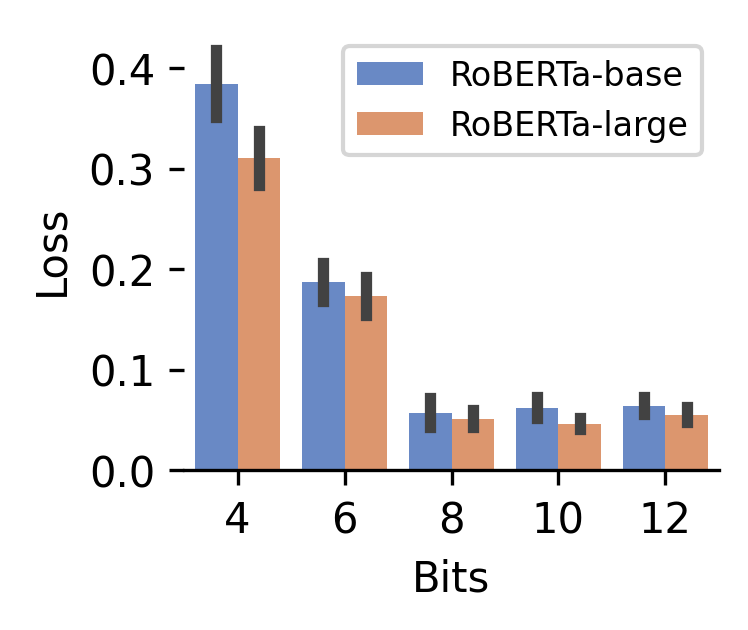}
	\end{minipage}
        \begin{minipage}{0.49\linewidth}
		\centering
        \hspace{-1cm}
		\includegraphics[width=0.95\linewidth]{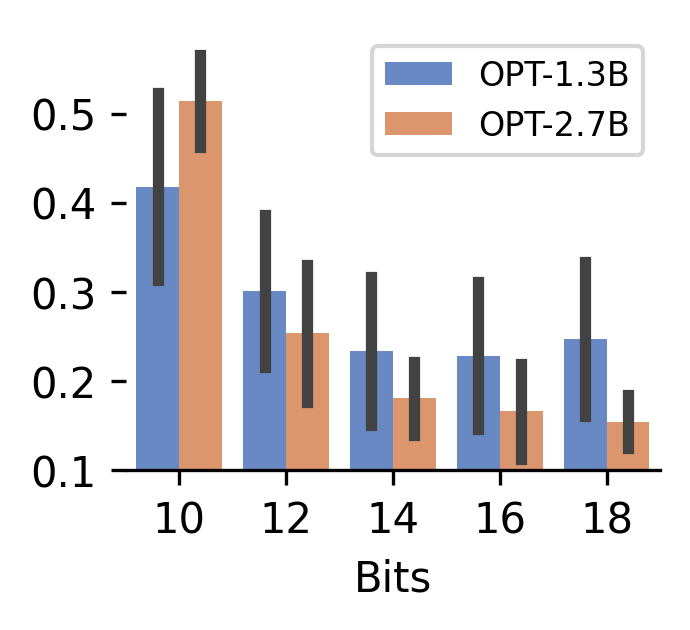}
	\end{minipage}
\vspace{-5pt}	
 \caption{Experiment on bit-width effectiveness. We train 1K and 10K steps for RoBERTa and OPT model, respectively.}
 \label{Figure4}
\vspace{-5pt}
\end{figure}


\noindent\textbf{How does the bit-width of RNGs affect performance?} 

In this subsection, we examine how the bit-width of RNGs in on-the-fly generation affects performance. Lower bit-width RNGs produce fewer unique perturbations, limiting optimization and preventing further loss reduction in later stages of training. Higher bit-width RNGs require more hardware resources and can increase the complexity of routing and timing closure. Thus, we need to measure the impact bit-width for the optimal trade-off, leveraging the least resources but maintaining high accuracy.

We trained LMs with varying RNG bit-widths and evaluated the final loss after extensive training steps. We observed that performance improves with higher bit-widths up to a threshold, beyond which no significant performance gains are shown. This threshold, termed the bottleneck bit-width, represents the point where further increases in the resolution of RNG no longer improve performance.


We trained LMs with varying RNG bit-widths and evaluated the final loss after extensive training steps. As shown in Figure~\ref{Figure4}, in medium-sized masked models like RoBERTa-base and RoBERTa-large, the training loss continues to decrease until the bit-width reaches 8 bits, further
increase has no noticeable performance improvement. This suggests that an 8-bit RNG is sufficient for the RoBERTa series models. For larger autoregressive models, OPT-1.3B and OPT-2.7B, the loss decreases noticeably until the bit-width reaches 14 bits, indicating that a 14-bit RNG is necessary to fully exploit the performance potential of the two OPT models.






\noindent\textbf{Hardware implementation analysis.}
Table~\ref{hardware} compares hardware utilization only for the random number generation part in ZO optimization. We choose the original MeZO as the baseline. Without using our PeZO, the baseline requires 1024 Gaussian random number generators to generate random numbers in parallel to match a typical tiling size of 1024 \cite{280896,10689464}. We consider the SOTA efficient GRNG design (TreeGRNG~\cite{crols2024treegrng}) to be used in the baseline.
As the results show, only the random number generation part (without considering the model inference) has already taken 48.6\% and 12.7\% overall LUTs and FFs of ZCU102 FPGA, and 4.474W overall power consumption.
This is an infeasible solution because of the huge hardware resource occupation, making it unavailable for the implementation of other functional modules.
With our PeZO framework, our pre-generation strategy only requires storing 4096 12-bit random numbers in BRAMs. We choose to split them into 8 BRAMs to alleviate the bandwidth pressure between the computation and the on-chip memory and enable the pipeline computation brought with the fine-tuning.
For on-the-fly generation, our method only requires 32 URNGs and a minimum of 8-bit or 14-bit requirements for the random numbers for RoBERTa and OPT models, respectively. Therefore, a total number of 32 LUTs, 449 FFs, or 512 FFs is needed.
During the implementation, it is observed that our optimized design can achieve a higher maximum operating frequency of 700 MHz, compared to the 500 MHz of the baseline design from the less congested floorplan and layout with less resource investment. This advantage can help increase the throughput for further processing. 
From the perspective of power consumption, compared with the SoTA solution to provide random numbers for ZO, our method saves 53\% and 86\% of power in pre-generation and on-the-fly generation scenarios, respectively. This achievement proves PeZO's ability to provide energy-efficient performance. 

Compared to MeZO, our method significantly reduces the required hardware resource utilization and power consumption, making ZO optimization a feasible solution for memory-saving on-device training. Moreover, it is worth noting that pre-generation and on-the-fly generation exhibit different emphases on hardware resource requirements, enabling designers to flexibly design based on their needs. 

\begin{table}[t]
\centering
\vspace{30pt}
\caption{Hardware implementation of random number generation part on ZCU102 FPGA. Pre-gen: pre-generation. O-t-F: on-the-fly generation.}
\setlength{\tabcolsep}{1.3pt}
\scalebox{0.8}{
\begin{tabular}{lccccc}
\hline
\textbf{Method}     & \textbf{LUTs}   & \textbf{FFs}   & \textbf{BRAMs} & \textbf{\begin{tabular}[c]{@{}c@{}}Power \\ (W)\end{tabular}} & \textbf{\begin{tabular}[c]{@{}c@{}}Max Freq\\ (MHz)\end{tabular}} \\ \toprule
ZCU102 available    & 274080          & 548160         & 150            & -                                                             &                                                                    \\ \midrule
MeZO~\cite{malladi2023fine}               & 133120 (48.6\%) & 69632 (12.7\%) & -              & 4.474                                                         & 500                                                                  \\ \midrule
Ours Pre-gen      & -               & 16             & 8              & 2.104                                                         & 700                                                                \\ 
Ours O-t-F (RoBERTa) & 32              & 449            & 1              & 0.608                                                         & 700                                                                \\ 
Ours O-t-F (OPT)     & 32              & 512            & 1              & 0.626                                                         & 700                                                                \\ \bottomrule
\end{tabular}
}
\label{hardware}
\vspace{-10pt}
\end{table}

\subsection{How to Use ZO Optimization for Hardware-friendly On-device Training - A Guideline}

In this section, we summarize the patterns we find through the extensive experimental results and present them in the form of a guideline to help fine-tune LMs via ZO on hardware platforms. Our guideline is presented as follows: 
\begin{itemize}[leftmargin=9pt]
    \item For the two random number generation strategies, pre-generation and on-the-fly generation. The former is more suitable for relatively difficult optimization tasks, while the latter is suitable for relatively simple ones.
    \item Instead of generating independent random numbers for each weight, adopting an effective reuse strategy offers a more hardware-friendly solution. Pre-generate as few as $2^{12}$ random numbers or implementing $2^5$ RNG, plus a reuse strategy, can produce competitive performance.
    \item Instead of directly implementing a high-bit RNG, using a low-bit RNG combined with RNG shifting to extend the cycle of randomness, would be a more hardware-efficient solution. An RNG with 8 to 14 bits, along with RNG shifting, is sufficient to achieve strong results.
\end{itemize}









\section{Related work and Discussion}

\subsection{Memory-efficient Fine-tuning for Fundamental Models}

Fundamental models, also known as pre-trained models, can be quickly transferred to downstream tasks through fine-tuning and have achieved state-of-the-art performance in a wide range of fields. However, the sheer size and complexity of these models, which often with billions of parameters, pose significant memory and computational challenges, especially for on-device training. This has motivated the development of memory-efficient fine-tuning methods, enabling researchers and practitioners to adapt models with non-industry-standard hardware. One of the most influential directions is parameter-efficient fine-tuning (PEFT), which updates only a small subset of model parameters while keeping the rest frozen. A prime example is Low-Rank Adaptation (LoRA)~\cite{hu2021lora, azizi2024lamda}, where low-rank matrices are injected into the model’s architecture to approximate parameter updates without modifying the original weights. Another promising strategy is sparse fine-tuning~\cite{liu2024sparse, zhang2023dynamic}, which selectively activates neurons or weights during training, drastically cutting memory and computation without hurting accuracy. Beyond PEFT, several memory-saving techniques at the system or training level have also proven effective, like Gradient checkpointing~\cite{feng2021optimal} and quantization-based methods~\cite{dettmers2022gpt3}.

An increasingly valuable direction is zeroth-order (ZO) optimization for model fine-tuning. Unlike backpropagation-based methods that rely on gradient computations, which are memory-intensive due to the need to store intermediate gradients and activations, ZO approaches optimize a model without computing explicit gradients, also, without complex backpropagation. Therefore, ZO optimization addresses the two primary challenges of intensive memory costs and complex computation, making it a promising solution for on-device training, especially for resource-limited edge devices like FPGAs and ASICs.

Though promising, the practical implementation difficulty for on-device ZO optimization has been overlooked by previous works, which is mainly caused by the requirement of numerous Gaussian random numbers for perturbation (detailed in Section~\ref{challenges}). Some straightforward methods could be a solution at first glance. For example, use Rademacher perturbations (i.e., -1, +1) instead of Gaussian for perturbation, which could be hardware-friendly, but the severe accuracy drop (shown in Table~\ref{table1}) makes it infeasible for fine-tuning large pre-trained models. Moreover, combining the PEFT method with ZO, therefore obviously reduces the needed number of random numbers, as weights needed to fine-tune are much less than full-parameter fine-tuning. Although some PEFT works claim that they only need to fine-tune 1\% or even 0.1\% weight to achieve competitive accuracy, the needed random number is still unacceptable for on-device training, considering the parameters of billions.

\subsection{Limitation and future work}

As this is the first work that explores the feasibility of ZO optimization on hardware, there are a few limitations of our work, which are discussed as follows.

First, our paper mainly focuses on testing fine-tuning language models for NLP tasks and does not include a broad range of application domains (e.g., computer vision tasks). This is because the current research in ZO methods is mainly targeted at the NLP tasks. Previous work finds that the data augmentation in NLP tasks (i.e., prompts) is crucial for ZO methods to achieve decent accuracy~\cite{malladi2023fine}. 
But the augmentation for other fields like computer vision has not been explored yet, which is mainly related to algorithm-level design and is orthogonal to our work. We leave it as future work.

Second, ZO method exhibits noisy convergence and requires more iterations compared to gradient-based methods to achieve comparable results. Therefore, in some cases, ZO method underperforms relative to gradient-based fine-tuning methods. There are some recent works~\cite{zhao2024second,tan2025harmony} targeted at addressing this issue. Note that our PeZO is compatible with different types of ZO methods, and it will be future work to integrate our framework with these methods for enhancing the performance of on-device ZO training.

\section{Conclusion}


In this paper, we propose PeZO, a novel perturbation-efficient ZO fine-tuning framework addressing an important but long-ignored problem when deploying ZO optimization for on-device training. With two random number reuse strategies and hardware-friendly adaptive modulus scaling, our method with extremely restricted resources for random number generation, can achieves highly competitive accuracy with the representative baseline, which with an ideal perturbation resource. Our method reduces the required LUTs and FFs for random number generation by 48.6\% and 12.7\%, and at maximum saves 86\% power consumption. Therefore, PeZO makes ZO a feasible paradigm for on-device training and also provides guidelines and valuable insights for future research.

    

\bibliographystyle{ACM-Reference-Format}
\bibliography{bib}

\end{document}